\documentclass[11pt,a4paper]{article}
\usepackage[hyperref]{acl2020}
\usepackage{times}
\usepackage{latexsym}
\usepackage{subcaption}
\usepackage{graphicx}
\usepackage{amssymb}
\usepackage{amsmath}
\usepackage{algorithm}
\usepackage{algorithmicx}
\usepackage{algpseudocode}
\usepackage{multirow}

\usepackage{microtype}

\aclfinalcopy 

\title{Capsule-Transformer for Neural Machine Translation}
\author{Sufeng Duan$^{1,2,3}$, Juncheng Cao$^{1,2,3}$, Hai Zhao$^{1,2,3}$\thanks{$^{*}$Corresponding author. This paper was partially supported by National Key Research and Development Program of China (No. 2017YFB0304100) and Key Projects of National Natural Science Foundation of China (No. U1836222 and No. 61733011).}  \\
	$^{1}$Department of Computer Science and Engineering, Shanghai Jiao Tong University \\
	$^{2}$Key Laboratory of Shanghai Education Commission for Intelligent Interaction \\ and Cognitive Engineering, Shanghai Jiao Tong University, Shanghai, China\\
	$^{3}$MoE Key Lab of Artificial Intelligence, AI Institute, Shanghai Jiao Tong University \\
{\tt 1140339019dsf@sjtu.edu.cn, caojuncheng@sjtu.edu.cn, zhaohai@cs.sjtu.edu.cn}}

\date{}

\begin{document}
\maketitle
\begin{abstract}
Transformer hugely benefits from its key design of the multi-head self-attention network (SAN), which extracts information from various perspectives through transforming the given input into different subspaces.
However, its simple linear transformation aggregation strategy may still potentially fail to fully capture deeper contextualized information.
In this paper, we thus propose the capsule-Transformer, which extends the linear transformation into a more general capsule routing algorithm by taking SAN as a special case of capsule network.
So that the resulted capsule-Transformer is capable of obtaining a better attention distribution representation of the input sequence via information aggregation among different heads and words.
Specifically, we see groups of attention weights in SAN as low layer capsules.
By applying the iterative capsule routing algorithm they can be further aggregated into high layer capsules which contain deeper contextualized information.
Experimental results on the widely-used machine translation datasets show our proposed capsule-Transformer outperforms strong Transformer baseline significantly.
\end{abstract}

\section{Introduction}
\begin{figure}[t]
    \centering
    \includegraphics[width=0.44\textwidth]{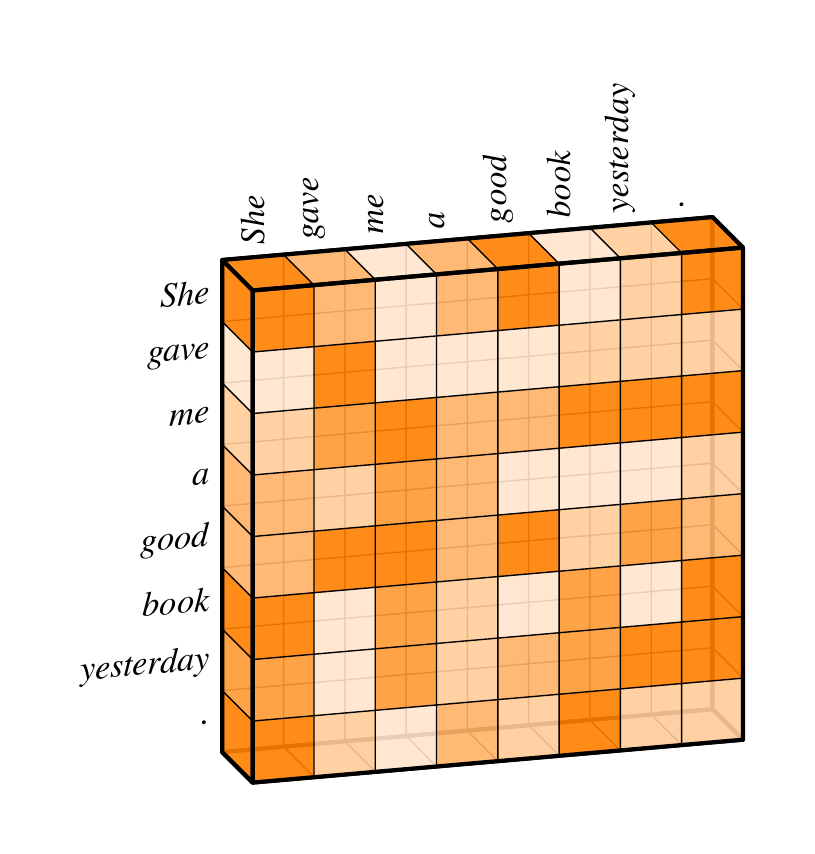}
    \caption{Demonstration of a pseudo self-attention weight matrix calculated from the sentence ``\textit{She gave me a good book yesterday.}'' on one head. Deeper color represents higher attention. It is reasonable to gather groups of attention weights (neurons) to compose various attention capsules.}
    \label{pic:demo}
\end{figure}
Witnessing the impressive results obtained in the field of machine translation \citep{bahdanau2014neural,luong-etal-2015-effective}, the implementation of attention mechanism and its variants quickly becomes a standard component in neural networks when facing the tasks such as document classification \citep{yang-etal-2016-hierarchical}, speech recognition \citep{chorowski2015attention} and many other natural language processing (NLP) applications, which help achieve promising performance compared to previous work. 
However, most of early work only implemented the attention mechanism on a recurrent neural network (RNN) architecture e.g. Long Short-Term Memory (LSTM) \citep{hochreiter1997long} and Gated Recurrent Unit (GRU) \citep{cho-etal-2014-learning} which have the problem of lacking the support of parallel computation, making it unpractical to build deep network.
In order to address the problem above, \citet{vaswani2017attention} proposed a novel self-attention network (SAN) architecture empowered by multi-head self-attention, which utilizes different heads to capture partial sentences information by projecting the input sequence into multiple distinct subspaces in parallel.
Although they only employed the simple linear transformations on the projection step, the impressive performance of the Transformer network still achieves a great success. 

Most existing work that focused on the improvement of multi-head attention mechanism mainly try to extract a more informative partial representation on each independent heads \citep{lin2017structured}.
\citet{li-etal-2019-information} proposed aggregating the output representations of multi-head attention.
\citet{dou2019dynamic} tried to dynamically aggregate information between the output representations from different encoder layers.
All these work concentrates mainly on the parts either ``before'' or ``after'' the step of multi-head SAN, which, as an important part of the whole Transformer model, should be paid more attention to.
To more empower the current Transformer, we thus propose constructing a more general and context-aware SAN, so that the model can learn deeper contextualized information of the input sequence, which could eventually be helpful in improving the model final performance.

In this paper, we propose the novel \textit{capsule-Transformer}, in which we implement a generalized SAN called \textit{Capsule Routing Self-Attention Network} which extends the linear transformation into a more general capsule routing algorithm \citep{sabour2017dynamic} by taking SAN as a special case of capsule network.
One of the biggest changes from capsule networking mechanism is altering the processing unit from scalar (single neuron) to capsule (group of neurons or vectors).
Inspired by the idea of such a capsule processing, we first similarly organize groups of attention weights calculated through self-attention into various capsules containing preliminary 
linguistic features, then we apply the routing algorithm on these capsules to obtain an output which can contain deeper contextualized information of the sequence.
Re-organizing the SAN in a capsule way, we extend the model to a more general form compared to the original SAN.

\section{Background}
\paragraph{Self-Attention}
The attention mechanism was first introduced into the machine translation models \citep{bahdanau2014neural,luong-etal-2015-effective} and has received well development and broad applications for its ability to effectively model the dependencies without regard to the distance between the input and output sequences. 
The Transformer is proposed in \citet{vaswani2017attention} empowered by multi-head attention mechanism which leverages multiple distinct transformation matrices, as different heads, to capture context information from various subspaces.
While the input and output sequences are the same for self-attention.

Formally, the multi-head self-attention mechanism models the attention calculation as a query operation with some specific keys.
Given the input sequence hidden states of query, key and value as $\{\mathbf{Q}, \mathbf{K}, \mathbf{V}\}$, where $\mathbf{Q} = \mathbf{K} = \mathbf{V} \in \mathbb{R}^{L \times d}$.
Here $d$ denotes the word embedding dimension, and $L$ is the length of the input sequence.
In multi-head attention, the $\mathbf{Q}$, $\mathbf{K}$ and $\mathbf{V}$ will be projected to $H$ different subspaces if there are $H$ heads in the model.
The transformation functions are all trainable linear matrices:
\begin{align}
    \mathbf{Q}_h, \mathbf{K}_h, \mathbf{V}_h = \mathbf{Q}\mathbf{W}^Q_h, \mathbf{K}\mathbf{W}^K_h, \mathbf{V}\mathbf{W}^V_h
\end{align}
where $\mathbf{Q}_h$, $\mathbf{K}_h$ and $\mathbf{V}_h$ are the projection of the original $\mathbf{Q}$, $\mathbf{K}$ and $\mathbf{V}$ on the $h^{th}$ subspace (head).
The size of each of the transformation matrices $\{\mathbf{W}^Q_h, \mathbf{W}^K_h, \mathbf{W}^V_h\}$ is $d \times d/H$.

An attention function $\textsc{Att}(\cdot)$ is applied on each head over the projected $\mathbf{Q}_h$ and $\mathbf{K}_h$.
The output on each head $\{\mathbf{O}_1, \ldots, \mathbf{O}_H\}$ is computed by combining the attention results with the value matrix $\mathbf{V}_h$ as:
\begin{align}
    \mathbf{O}_h &= \textsc{Att}(\mathbf{Q}_h, \mathbf{K}_h)\mathbf{V}_h \\
    \mathbf{O} &= [\mathbf{O}_1, \ldots, \mathbf{O}_H]
\end{align}
where $\mathbf{O}_h \in \mathbb{R}^{L \times d/H}$, and $\mathbf{O} \in \mathbb{R}^{L \times d}$ is the concatenation of $H$ partial outputs from all the heads.

In this paper, we adopt the $\textsc{Att}(\cdot)$ with scaled dot-product attention function \citep{luong-etal-2015-effective} which is faster and more suitable for the parallel computation compared to the additive attention \citep{vaswani2017attention}:
\begin{align}
    \textsc{Att}(\mathbf{Q}_h, \mathbf{K}_h) &= softmax(\mathbf{E}_h) \\
    \begin{split}
        \label{eq:attn}
        \mathbf{E}_h &= \frac{\mathbf{Q}_h \mathbf{K}_h^T}{\sqrt{d_k}} \\
        &= [\mathbf{e}_{1,h}^T, \ldots, \mathbf{e}_{L,h}^T]^T
    \end{split}
\end{align}
where $\mathbf{e}_{l,h} \in \mathbb{R}^L$ is the computed attention vector of the $l^{th}$ token of the input sequence on the $h^{th}$ head.

From Eq.~(\ref{eq:attn}), we are aware that the attention vector $\mathbf{e}_{l,h}$ containing the crucial attentive clues is important to compose the final output $\mathbf{O}$. We view $\mathbf{e}_{l,h}$ as an entity basis to conduct a more general self-attention mechanism.

\paragraph{Capsule Network}
Instead of applying operations in individual neurons as common neural network,
    capsule network takes 
    a capsule as the basic processing unit which consists of a group of neurons.
    The way of connection between two capsule layers is also different from conventional neuron based networks.
    Formally, for capsule $\mathbf{C}^{l}_i$ in layer $l$, it will generate a vote vector $\mathbf{V}_{i \rightarrow j}$ to determine to what extent itself belongs to the capsule $\mathbf{C}^{l+1}_j$ in layer $l+1$.
    Similar to the fully connected layer, the $\mathbf{C}^{l+1}_j$ thus will be composed of all the vote vectors $\mathbf{V}_{* \rightarrow j}$ generated from the layer $l$.

\section{Model Architecture}
\subsection{Vertical and Horizontal Capsules}
\begin{figure}[tbp]
    \centering
    \includegraphics[width=0.44\textwidth]{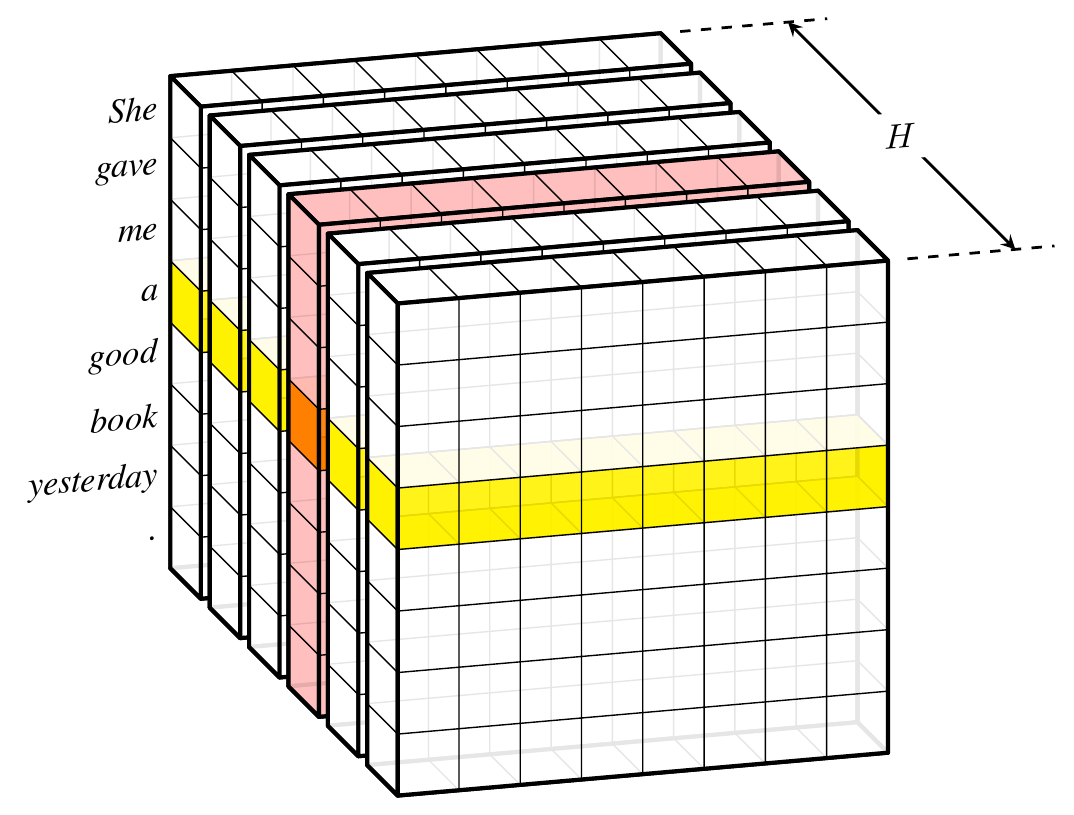}
    \caption{Vertical and horizontal capsules. The red block represents one of the vertical capsules and the yellow block represents one of the horizontal capsules. Their overlapping attention vector is marked with orange.}
    \label{pic:caps}
\end{figure}
Capsule network was proposed in \citet{sabour2017dynamic} on the field of computer vision which changed the conventional way of data flow in neural networks.
Concretely speaking, the capsule network views a group of neurons (scalars) which captures the parameters of some specific feature as a capsule entity.
In computer vision, that kind of feature could be the detection of eyes, nose or mouth when doing a face recognition task.
Through the advanced capsule routing algorithm proposed in their work, the low level feature capsules can be aggregated to form the high level capsules which may represent some more abstract features such as a human face or a left arm.

When it turns to multi-head SAN in NLP tasks, it is coincidentally lucky that the calculated attention weights have already been organized into multiple separate groups.
Considering that these groups of weights represent partial information of the input sequence from different perspectives or subspaces, they could be naturally treated as capsules. Then, it is intuitively for us to extend the original linear transformation aggregation as the capsule routing algorithm on these capsules so that a better attention distribution representation can be obtained via careful capsule routing.

In this paper, we organize the attention weights into two types of capsule: 1) capsule that contains all the attention weights on one of the $H$ heads; 2) capsule that contains attention weights of one of the $L$ tokens on all the heads.
As shown in Figure \ref{pic:caps}, we name the head-wise capsules with \textit{vertical capsules} $\mathbf{C}^{\updownarrow}_h$ and the token-wise capsules with \textit{horizontal capsules} $\mathbf{C}^{\leftrightarrow}_l$ according to their placement way in the attention weights cube respectively.

\subsection{Overview}
\begin{figure}[tbp]
    \centering
    \includegraphics[width=0.4\textwidth]{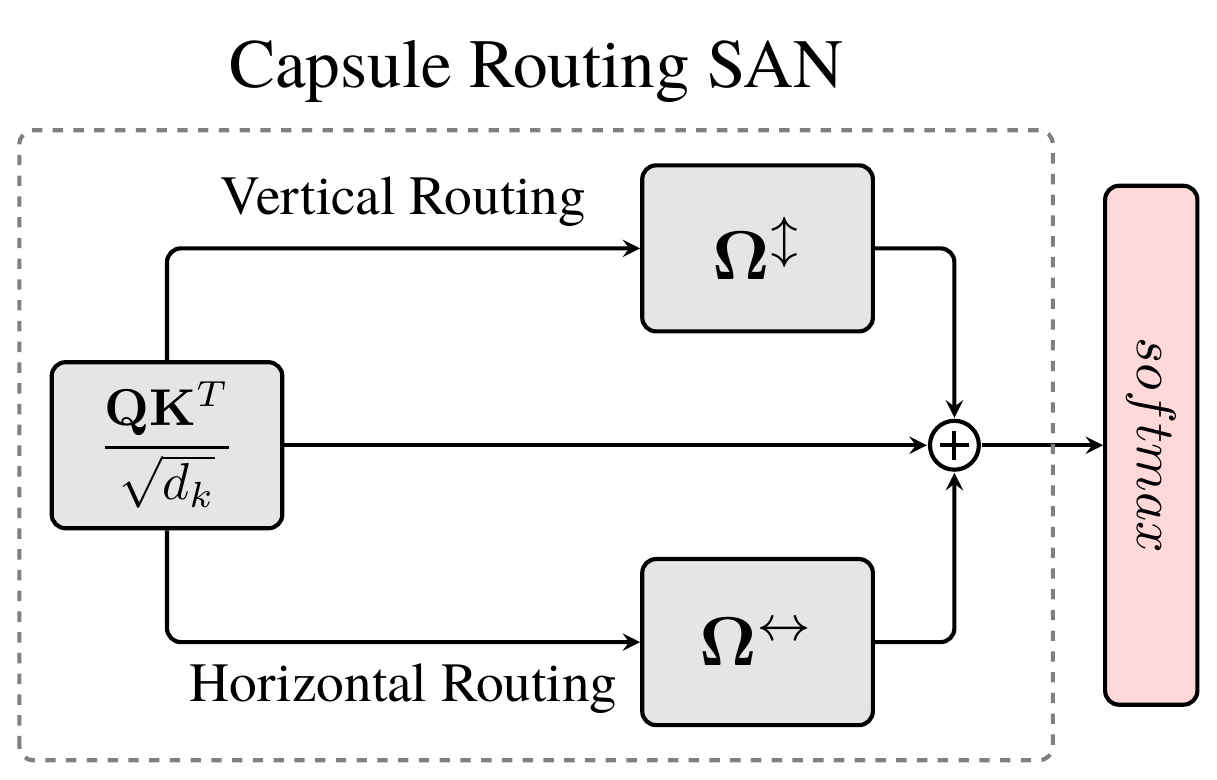}
    \caption{The overview of capsule routing SAN.}
    \label{pic:overview}
\end{figure}
A simple architecture of our capsule routing SAN in capsule-Transformer is shown in Figure \ref{pic:overview}, and the dashed part is the main difference between our model and the vanilla one.
Our generalized SAN inspired by capsule routing is composed of two components: the vertical routing part and the horizontal routing part.
Each part will do a capsule routing on the attention weight matrices calculated through scaled dot-product and obtain the corresponding output capsules: the vertical output capsule $\mathbf{\Omega}^{\updownarrow}$ and the horizontal output capsule $\mathbf{\Omega}^{\leftrightarrow}$.
Both $\mathbf{\Omega}^{\updownarrow}$ and $\mathbf{\Omega}^{\leftrightarrow}$ have the same size as the input attention cube.
Before the softmax, we add these two output capsules to the original attention matrices so that every token in the input sequence can get a better contextualized attention distribution representation on each head.

\subsection{Routing Algorithm}
Following \citet{sabour2017dynamic}, we adopt the
\textit{Dynamic Routing} algorithm as shown in Algorithm \ref{alg:dynamic-routing}. 
\begin{algorithm}[t]
    \caption{Dynamic Routing}
    \label{alg:dynamic-routing}
    \begin{algorithmic}
        \Require $M \times N$ vote vectors $\mathbf{V}_{m \rightarrow n}$, iteration times $T$
        \Ensure $N$ output capsules $\mathbf{\Omega}_n$, vote weights $\mathbf{B}$
        \Function {Routing}{$\mathbf{V}, T$}
            \State $\forall~\mathbf{V}_{m \rightarrow n}$: $B_{m \rightarrow n} \gets 0$
            \For {$T$ iterations}
                \State $\forall~(m,n)$: $R_{m \rightarrow n} \gets softmax(B_{m \rightarrow *})$
                \State $\forall~\mathbf{\Omega}_n$: compute $\mathbf{\Omega}_n$ by Eq.~(\ref{eq:squashing})
                \State $\forall~(m,n)$: $B_{m \rightarrow n} \mathrel{+}= \mathbf{\Omega}_n \cdot \mathbf{V}_{m \rightarrow n}$
            \EndFor
            \State \Return{$\mathbf{\Omega}$, $\mathbf{B}$}
        \EndFunction
    \end{algorithmic}
\end{algorithm}

Formally, the dynamic routing algorithm is applied between two capsule layers which are called \textit{input capsule layer} and \textit{output capsule layer}.
Suppose there are $M$ and $N$ capsules in the input and output capsule layer respectively, then each of the $M$ input capsules should generates $N$ \textit{vote vectors} associated with the corresponding output capsules.
The vote vectors generated from the $m^{th}$ input capsule and associated with the $n^{th}$ output capsule will take the job of measuring the belonging relationship between those two capsules.

For each vote vector $\mathbf{V}_{m \rightarrow n}$, a weight value $R_{m \rightarrow n}$ will be dynamically assigned on it.
The output capsule $\mathbf{\Omega}_n$ is computed by a scaled weighted sum of all its associated vote vectors:
\begin{align}
    \label{eq:squashing}
    \mathbf{\Omega}_n &= \frac{\|\mathbf{S}_n\|^2}{1 + \|\mathbf{S}_n\|^2} \frac{\mathbf{S}_n}{\|\mathbf{S}_n\|} \\
    \mathbf{S}_n &= \sum_{m=1}^M R_{m \rightarrow n}\mathbf{V}_{m \rightarrow n}
\end{align}
where Eq.~(\ref{eq:squashing}) computing $\mathbf{\Omega}_n$ is a non-linear ``squashing'' function used by \citet{sabour2017dynamic}, aiming to measure the existence probability of the aggregated capsules via its length.
The weight value $R_{m \rightarrow n}$ is dynamically updated by applying the softmax function on the accumulated sum ${B_{m \rightarrow *}}$ in each iteration, and the accumulated number is determined by the scalar product of $\mathbf{\Omega}_n$ and $\mathbf{V}_{m \rightarrow n}$.

\subsection{Capsule Routing Self-Attention Network}
As mentioned above, in our capsule routing SAN, we reorganize the attention weight matrices into head-wise vertical capsules $\mathbf{C}^{\updownarrow}_h$ and token-wise horizontal capsules $\mathbf{C}^{\leftrightarrow}_l$.
Therefore the rest problem here for us is how to generate various vote vectors based on these two types of capsules.

Most existing work simply applied multiple linear transformations on the capsule to generate vote vectors \citep{sabour2017dynamic,li-etal-2019-information,dou2019dynamic}.
However, such linear processing method cannot be used here for the capsule sizes are not fixed.
Therefore here we do not use the method of applying trainable parameterized transformation functions on capsules to generate vote vectors.
Instead, we leverage the original attention vector $\mathbf{e}_{l,h}$ to do the voting job.
Since dynamic routing is also an nonparametric algorithm, we could barely introduce no new parameters to our capsule-Transformer, in which 
the way of vote vector generation will less the least to hurt the model performance.

\subsubsection{Vertical Routing}
\begin{figure*}[t]
    \centering
    \includegraphics[width=0.7\textwidth]{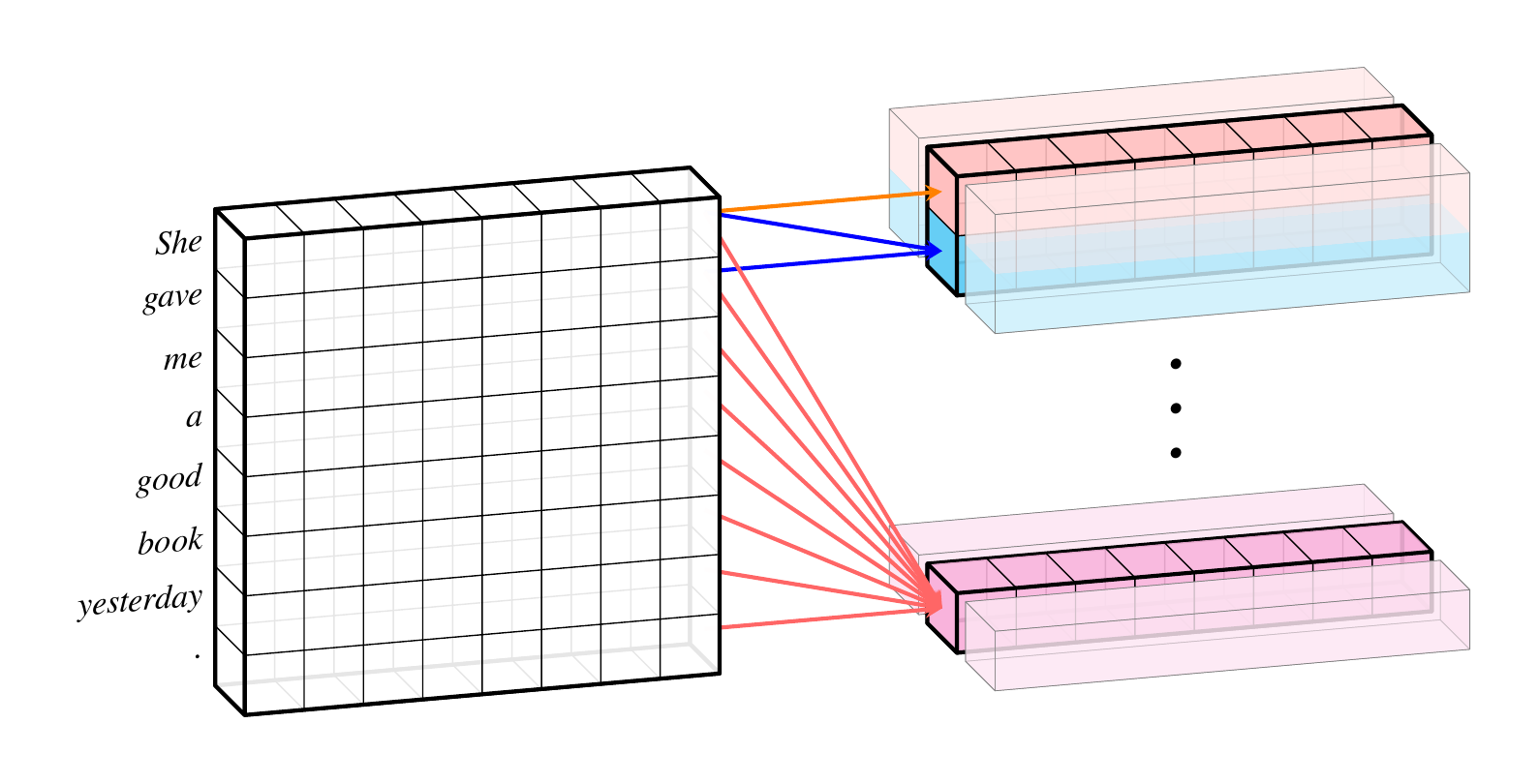}
    \caption{Positional Routing.}
    \label{pic:pos}
\end{figure*}
\begin{algorithm}[t]
    \caption{Vertical Routing}
    \label{alg:ver-routing}
    \begin{algorithmic}
        \Require $H \times L$ vote vectors $\mathbf{V}^{\updownarrow}_{h \rightarrow l}$, iteration times $T$
        \Ensure Vertical output capsule $\mathbf{\Omega}^{\updownarrow}$
        \Function{VerticalRouting}{$\{\mathbf{V}^{\updownarrow}_{h \rightarrow l}\}, T$}
            \State $\mathbf{\Omega}^{\updownarrow}_l, \mathbf{B} \gets$ \Call{Routing}{$\{\mathbf{V}^{\updownarrow}_{h \rightarrow l}\}, T$}
            \State $\widetilde{\mathbf{\Omega}}^{\updownarrow} \gets Concat(\{\mathbf{\Omega}^{\updownarrow}_l\})$
            \State $\mathbf{\Lambda} \gets$ computed by Eq.~(\ref{eq:lambda})
            \State $\lambda_h \gets softmax(\mathbf{\Lambda})$
            \State $\mathbf{\Omega}^{\updownarrow} \gets Concat(\{\lambda_h \widetilde{\mathbf{\Omega}}^{\updownarrow}\})$
            \State \Return{$\mathbf{\Omega}^{\updownarrow}$}
        \EndFunction
    \end{algorithmic}
\end{algorithm}

In vertical routing, we view the attention weight cube head-wise so that in an $H$ head Transformer model we could get $H$ vertical capsules $\mathbf{C}^{\updownarrow}_h \in \mathbb{R}^{L \times K}$, where $K$ is the length of the attention vector.
And in self-attention, $K = L$.
We could split the vertical capsule $\mathbf{C}^{\updownarrow}_h$ into $L$ vote vectors:
\begin{align}
    \mathbf{V}^{\updownarrow}_{h \rightarrow l} = \mathbf{e}_{l,h}
\end{align}

By applying the dynamic routing algorithm, we can calculate the vertical output capsule:
\begin{align}
    \{\mathbf{\Omega}^{\updownarrow}_l\} = \textsc{Routing}(\{\mathbf{V}^{\updownarrow}_{h \rightarrow l}\},T) \in \mathbb{R}^K
\end{align}

Simply adding the $L$ output capsules to all the $H$ heads may ignore an obvious fact 
that each head makes effects in different degrees in the formation of the output capsules, so that every heads may ``absorb''  the output capsule also differently.
In the meantime, it has been found that for a deep layered Transformer, it has a hierarchical pattern of information capturing \citep{raganato-tiedemann-2018-analysis,peters-etal-2018-deep}.
Taking both of the above issues into consideration, 
we use a trainable linear matrix $\mathbf{W}^{\updownarrow} \in \mathbb{R}^{H \times H}$ and bias $b$ in each layer to measure the extent of the output capsule acceptance:
\begin{align}
    \mathbf{\Omega}^{\updownarrow} &= softmax(\mathbf{\Lambda})
    \begin{bmatrix}
        \mathbf{\Omega}^{\updownarrow}_1 \\
        \vdots \\
        \mathbf{\Omega}^{\updownarrow}_L
    \end{bmatrix} \\
    \label{eq:lambda}
    \mathbf{\Lambda} &= \mathbf{W}^{\updownarrow}\left[\sum_{l=1}^L B_{1 \rightarrow l}, \ldots, \sum_{l=1}^L B_{H \rightarrow l}\right] + b
\end{align}
where $B_{h \rightarrow l}$ are the vote weights calculated in the last iteration in the routing.

\subsubsection{Horizontal Routing}
\begin{algorithm}[t]
    \caption{Horizontal Routing}
    \label{alg:hor-routing}
    \begin{algorithmic}
        \Require $L \times H$ vote vectors $\mathbf{V}^{\leftrightarrow}_{l \rightarrow h}$, iteration times $T$
        \Ensure Horizontal output capsule $\mathbf{\Omega}^{\leftrightarrow}$
        \Function{HorizontalRouting}{$\{\mathbf{V}^{\leftrightarrow}_{l \rightarrow h}\}, T$}
            \For {$l=1 \to L$}
                \State $\mathbf{\Omega}^{\leftrightarrow}_l \gets$ \Call{Routing}{$\{\mathbf{V}^{\leftrightarrow}_{t \rightarrow h}\}, T$}, $t \le l$
            \EndFor
            \State $\mathbf{\Omega}^{\leftrightarrow} \gets Concat(\{\mathbf{\Omega}^{\leftrightarrow}_l\})$
            \State \Return{$\mathbf{\Omega}^{\leftrightarrow}$}
        \EndFunction
    \end{algorithmic}
\end{algorithm}
Different from vertical routing, we split the attention weight cube token-wise in the horizontal routing part.
So for an $L$-word sequence, we totally have $L$ horizontal capsules $\mathbf{C}^{\leftrightarrow}_l \in \mathbb{R}^{H \times K}$, in which each of the $L$ capsules will therefore generate $H$ vote vectors:
\begin{align}
    \mathbf{V}^{\leftrightarrow}_{l \rightarrow h} = \mathbf{e}_{l,h}
\end{align}

It is worth noting that there is one fundamental difference between the vertical and horizontal capsules that the former is order independent while the latter is not.
Such an essential difference indicates that simply applying the routing algorithm to all the horizontal capsules might fail to capture positional relationship information among tokens.
Meanwhile, the number of capsules in input layer varies according to the length of sequence, which makes it impossible to implement a linear transformation method mentioned above.
Therefore we design a novel positional routing method to leverage this salient information.

\paragraph{Positional Routing}
Rather than applying the routing algorithm all at once, we here for each horizontal capsule do a partial routing to implicitly encode the sequential information into the output capsules.

As shown in Figure \ref{pic:pos}, for an $L$-word input sequence, there is a need of totally $L$ times partial routing.
For the $l^{th}$ partial routing, only the top $l$ horizontal capsules of the attention cube are involved in the aggregation, which means we only use the information of the first $l$ tokens. 
More concretely speaking, for capsule $\mathbf{C}^{\leftrightarrow}_l$, its output capsule is computed by routing all the capsule $\mathbf{C}^{\leftrightarrow}_t$, where $t \le l$:
\begin{align}
    \mathbf{\Omega}^{\leftrightarrow} &= \left[\mathbf{\Omega}^{\leftrightarrow}_1, \ldots, \mathbf{\Omega}^{\leftrightarrow}_L\right] \\
    \mathbf{\Omega}^{\leftrightarrow}_l &= \textsc{Routing}(\{\mathbf{V}^{\leftrightarrow}_{t \rightarrow h}\},T)
\end{align}
where $\mathbf{\Omega}^{\leftrightarrow}_l \in \mathbb{R}^{H \times K}$.

\subsubsection{Masked Routing in Decoder}
In the vanilla Transformer model, all the encoder and decoder layers apply a multi-head SAN sub-layer \citep{vaswani2017attention}.
One small modification of adding a forward mask is made in decoder stack to prevent from extracting information from the non-predicted tokens.
Similar to such a treatment, we also use a forward mask on the attention weight cube on each head before the routing step.
Meanwhile we remove the vertical routing part since the information among different tokens will still be allowed to be exchanged in the softmax step of each iteration in the routing.

\section{Experiment}
\begin{table*}[t]
    \centering
    \setlength{\tabcolsep}{2mm}{
    \begin{tabular}{l|l||c|c}
        \textbf{System} & \textbf{Architecture} & \textbf{Zh-En} & \textbf{En-De} \\
        \hline
        \hline
        \multicolumn{4}{c}{\textit{Existing NMT Systems}} \\
        \hline
        \citet{wu2016google}
        & RNN with 8 layers & - & 26.30 \\
        \citet{gehring2017convolutional}
        & CNN with 15 layers & - & 26.36 \\
        \hline
        \multirow{2}*{\citet{vaswani2017attention}}
        & Transformer-\textit{Base} & - & 27.30 \\
        & Transformer-\textit{Big} & - & 28.40 \\
        \hline
        \citet{hassan2018achieving}
        & Transformer-\textit{Big} & 24.20 & - \\
        \hline
        \multirow{2}*{\citet{li-etal-2019-information}}
        & Transformer-\textit{Base} + Effective Aggregation & 24.68 & 27.98 \\
        & Transformer-\textit{Big} + Effective Aggregation & 25.00 & 28.96 \\
        \hline
        \hline
        \multicolumn{4}{c}{\textit{Our NMT Systems}} \\
        \hline
        \multirow{4}*{\textit{this work}}
        & Transformer-\textit{Base} & 24.28 & 27.43 \\
        & capsule-Transformer-\textit{Base} & 25.02 & 28.04 \\
        \cline{2-4}
        & Transformer-\textit{Big} & 24.71 & 28.42 \\
        & capsule-Transformer-\textit{Big} & 25.14 & 28.71
    \end{tabular}}
    \caption{Comparing with existing NMT systems on WMT17 Chinese-to-English (Zh-En) and WMT14 English-to-German (En-De) tasks.}
    \label{tab:result}
\end{table*}
\subsection{Setup}
Our proposed model is evaluated on the widely-used WMT17 Chinese-to-English (Zh-En) and WMT14 English-to-German (En-De) datasets.
These two datasets consist of total 20.6M and 4.6M sentence pairs, respectively.
For Zh-En task, we found it would be helpful for reducing the vocabulary size without hurting the model performance when we only keep the sentence pairs whose length is less than 50 during the training and validation.
We use the newsdev2017 and newstest2017 as the validation set and test set through the training.
While for En-De task, we use all the sentence pairs for our model training.
For model validation and test, we use the newstest2013 as the validation set and newstest2014 is used as the test set.
To further decrease the vocabulary size, we employ byte-pair encoding (BPE) \citep{sennrich2016neural} on the training datasets and set the merge operations as 32K for both the WMT17 and WMT14 corpora.

Our proposed capsule-Transformer is implemented on the Transformer architecture \citep{vaswani2017attention}.
For the configuration of the hyper-parameters on both \textit{Base} ans \textit{Big} model, we follow their setup to train our baseline model on Zh-En and En-De tasks.
The Transformer-\textit{Base} and \textit{Big} model differ at the word embedding size (512 vs. 1024), the count of attention heads (8 vs. 16) and the dimensionality of feed-forward network (2048 vs. 4096).
For \textit{Big} model, to prevent from over-fitting we set the dropout rate as 0.3 compared to 0.1 of that of the \textit{Base} model.
For \textit{Base} model, we set the batch size up to no more than 2048 tokens and the gradient will accumulate for 12 times before the back-propagation.
For \textit{Big} model, those parameters are set as 1024 tokens per batch and 24 times for gradient accumulation.
The framework we use to implement both the baseline and our capsule-Transformer is OpenNMT-py \citep{klein-etal-2017-opennmt}.
We choose the case-sensitive 4-gram BLEU score \citep{bleu} as the metric to evaluate the performance of our models and compare it with that of the existing models.
We train our \textit{Base} and \textit{Big} models on 2 and 3 NVIDIA GeForce GTX 1080Ti GPUs, respectively.

\subsection{Main Results}
Table \ref{tab:result} lists the main results on both the WMT17 Chinese-to-English (Zh-En) and WMT14 English-to-German (En-De) datasets.
As shown in the table, our capsule-Transformer model consistently improves the performance across both language pairs and model variations, which shows the effectiveness and generalization ability of our approach.
For WMT17 Zh-En task, our model outperforms all the models listed above, especially only the capsule-Transformer-\textit{Base} model could achieve a score higher even than the other \textit{Big} model.
For WMT14 En-De task, our model outperforms the corresponding baseline while inferior to the \textit{Big} model proposed by \citet{li-etal-2019-information}.
Considering their model introduces over 33M new parameters (while for our \textit{Big} model, this number is 1.6K) and uses a much larger batch size than ours (4096 vs. 1024) in the training process, it is reasonable for us to believe that our model would achieve a more promising score if the condition was the same.

\subsection{Analysis}
We conduct extensive analysis experiments on our capsule-Transformer to better evaluate the effects of each model component.
All the results below are produced with the Transformer-\textit{Base} model setup on WMT17 Zh-En task.
\begin{table*}[t]
    \begin{minipage}[t]{0.33\textwidth}
        \centering
        \begin{tabular}{l|c|c|c}
            \textbf{\#} & \textit{Enc} & \textit{Dec} & \textbf{BLEU} \\
            \hline
            \hline
            1 & - & - & 24.28 \\
            \hline
            2 & \checkmark & - & 24.87 \\
            3 & - & \checkmark & 24.65 \\
            4 & \checkmark & \checkmark & 25.02
        \end{tabular}
        \caption{Effect in encoder and decoder.}
        \label{tab:enc-dec}
    \end{minipage}
    \begin{minipage}[t]{0.33\textwidth}
        \centering
        \begin{tabular}{l|c|c|c}
            \textbf{\#} & \textit{Ver} & \textit{Hor} & \textbf{BLEU} \\
            \hline
            \hline
            1 & - & - & 24.28 \\
            \hline
            2 & \checkmark & - & 24.74 \\
            3 & - & \checkmark & 24.76 \\
            4 & \checkmark & \checkmark & 24.87
        \end{tabular}
        \caption{Effect of routing parts.}
        \label{tab:ver-hor}
    \end{minipage}
    \begin{minipage}[t]{0.33\textwidth}
        \centering
        \begin{tabular}{l|c|c}
            \textbf{\#} & \textit{Layers} & \textbf{BLEU} \\
            \hline
            \hline
            1 & - & 24.28 \\
            \hline
            2 & 1-3 & 24.64 \\
            3 & 4-6 & 24.48 \\
            4 & 1-6 & 24.87
        \end{tabular}
        \caption{Effect on different layers.}
        \label{tab:low-high}
    \end{minipage}
\end{table*}
\paragraph{Effect on Transformer Componets}
To evaluate the effect of capsule routing SAN in encoder and decoder, we perform an ablation study.
As shown in Table \ref{tab:enc-dec}, both encoder and decoder benefit from our capsule routing SAN.
Especially the modified decoder still outperforms the baseline even we have removed the vertical routing part, which demonstrates the effectiveness of our model.
The row 4 proves the complementarity of the encoder and decoder with capsule routing SAN.

\paragraph{Effect of Different Routing Parts}
To compare importance of vertical and horizontal routing parts in capsule routing SAN, we evaluate models by removing either of the two from encoder.
As shown in Table \ref{tab:ver-hor}, both vertical and horizontal routing help enhance the model.
Meanwhile the two routing parts achieve nearly the same score, which shows that it is meaningful to re-organize the attention cube in these two separate perspectives.

\paragraph{Effect on Different Layers}
Since the deep layered Transformer is found having a hierarchical pattern of captured information \citep{raganato-tiedemann-2018-analysis,peters-etal-2018-deep}, it is necessary to explore the working pattern of our capsule routing SAN on different layers.
As shown in Table \ref{tab:low-high}, although our approach improves the performance on both higher and lower layers, it works better on the lower layers.

\paragraph{Attention Visualization}
\begin{figure}[t]
    \centering
    \begin{subfigure}[t]{0.47\textwidth}
        \includegraphics[width=1\linewidth]{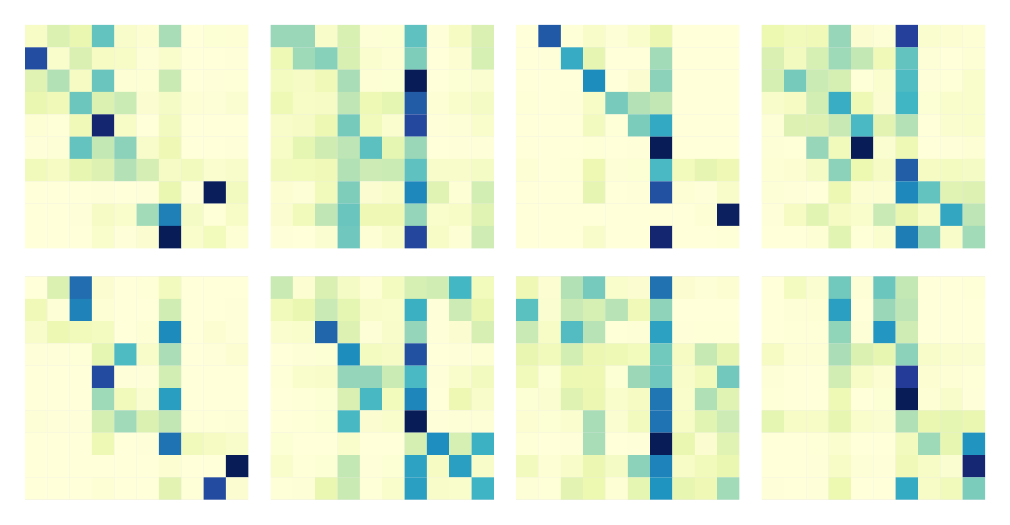}
        \caption{Transformer}
    \end{subfigure}
    \begin{subfigure}[t]{0.47\textwidth}
        \includegraphics[width=1\linewidth]{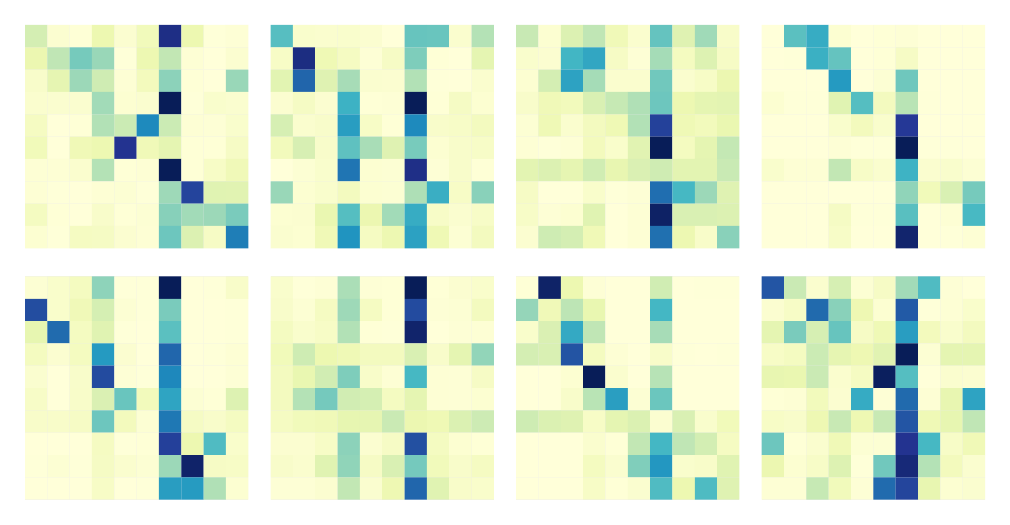}
        \caption{Capsule-Transformer}
    \end{subfigure}
    \caption{Attention visualization of SAN from vanilla and capsule-Transformer.}
    \label{pic:heatmap}
\end{figure}
To better understand the ability of our model in obtaining deep contextualized information, we randomly sample the sentence and visualize the attention weights of each head from the top encoder layer.
As shown in Figure \ref{pic:heatmap}, compared to the attention distribution of vanilla Transformer, it is clear that our model can capture more contextualized features on each head.

\section{Related Work}
Attention mechanism has become a standard component in nowadays neural machine translation models since it was first introduced by \citet{bahdanau2014neural}, in which an additive attention was implemented.
Later of that, \citet{luong-etal-2015-effective} applied a new attention method using dot-product, which is also inherited by \citet{vaswani2017attention} in their impressive work on Transformer.
Although Transformer model has advanced the state-of-the-art on various tasks of NLP, its ``over simple'' structure design implies that its potential capability might have not been fully exploited.

Adding extra information to Transformer can be the most intuitive way to enhance the model performance.
To alleviate the problem of lacking the consideration of positional information, \citet{shaw-etal-2018-self} extended the SAN with the incorporation of relative positional information.
\citet{shen2018disan} applied a directional mask method to encode the ``forward'', ``backward'' and ``local'' information in to the SAN.
So is the work done by \citet{cui-etal-2019-mixed}.
To model the localness of the sentences, \citet{yang-etal-2018-modeling} added a Gaussian bias to the attention weight vectors.
\citet{xiao-etal-2019-lattice} presented a lattice-based Transformer which integrates flexible segmentations into the encoder and the SAN.

However, nearly all these existing enhancement work can be viewed as adding hand-crafted features, implicitly or explicitly, to the SAN, which to some extent, might risk in losing the generalization ability.
Another choice is to encode the additional information through extra networks connected with the Transformer model.
In this way the model might learn deeper features.
To introduce a better ability to model the recurrence, \citet{hao-etal-2019-modeling} add the additional RNN encoder whose output embedding will be incorporated with the original one.
\citet{wang-etal-2019-tree} add a constraint from an extra bidirectional Transformer encoder to lead the attention heads to follow tree structures.

While networks with different structure may have unpredictable effects when they are incorporated together.
Some researchers focus on strengthening the Transformer via modifying the original structure.
\citet{yang2019context} leveraged deep and global context information to calculate a better attention between tokens.
\citet{guo-etal-2019-star} proposed a novel design of the SAN which simplifies the connection among each attention head.
\citet{yang-etal-2019-convolutional} used a CNN-like structure in SAN to model the localness by restricting the context size.
All these studies focus on the improvement of SAN due to its central role in Transformer modeling.
However, they still all view the elements in SAN as single scalars rather than in a more contextualized way.
Different from all the existing work, in this paper, we instead consider a generalized SAN design by inheriting the idea from capsule networking and taking SAN as a special case of capsule network. So that at last we can extend the vanilla Transformer model into a more generalized form.

\paragraph{Capsule Networks for NLP}
Capsule network was introduced to NLP to do the classification task \citep{yang-etal-2018-investigating,chen-qian-2019-transfer} for its capsule clustering mechanism can be simply used without much modification.
Its information aggregation mechanism was also utilized to encode the input sequence with fixed size \citep{gong2018information}, obtaining a better output hidden states of encoder by aggregating information from different layers \citep{dou2019dynamic}.
For SAN, \citet{li-etal-2019-information} applied the routing algorithm on the concatenated output representation while not change the SAN structure.
Recently, \citet{liu-etal-2019-transformer} combined the Transformer and capsule network for the stock movements prediction.
However, they only stacked Transformer encoders in a capsule way rather than deeply integrating the two networks.

Different from all the impressive improvement over the original Transformer, we adopt a deep architecture revision by generalizing the self-attention mechanism which empowers the Transformer most into a sort of capsule routing processing. Especially, we nearly introduce no new parameters for such a model design improvement, which keeps the merit of simple-idea-inspiring of the original Transformer.

\section{Conclusion}
In this paper, we propose the capsule-Transformer, which extends the linear transformation of self-attention in the vanilla Transformer into a more general capsule routing algorithm by taking SAN as a special case of capsule network.
So that the resulted capsule-Transformer is capable of obtaining a better attention distribution representation of the input sequence via information aggregation among different heads and words.
We verify the proposed capsule-Transformer only in the task of neural machine translation though, which already shows its superiority over the strong Transformer baseline, the proposed model architecture design has potentially a broad application prospect for various NLP tasks as the Transformer.

\bibliography{acl2020}

\begin{thebibliography}{33}
\expandafter\ifx\csname natexlab\endcsname\relax\def\natexlab#1{#1}\fi

\bibitem[{Bahdanau et~al.(2015)Bahdanau, Cho, and Bengio}]{bahdanau2014neural}
Dzmitry Bahdanau, Kyunghyun Cho, and Yoshua Bengio. 2015.
\newblock \href {https://arxiv.org/pdf/1409.0473.pdf} {Neural machine
  translation by jointly learning to align and translate}.
\newblock In \emph{Proceedings of the 3rd International Conference on Learning
  Representations (ICLR 2015)}.

\bibitem[{Chen and Qian(2019)}]{chen-qian-2019-transfer}
Zhuang Chen and Tieyun Qian. 2019.
\newblock \href {https://doi.org/10.18653/v1/P19-1052} {Transfer capsule
  network for aspect level sentiment classification}.
\newblock In \emph{Proceedings of the 57th Annual Meeting of the Association
  for Computational Linguistics (ACL 2019)}, pages 547--556.

\bibitem[{Cho et~al.(2014)Cho, van Merri{\"e}nboer, Gulcehre, Bahdanau,
  Bougares, Schwenk, and Bengio}]{cho-etal-2014-learning}
Kyunghyun Cho, Bart van Merri{\"e}nboer, Caglar Gulcehre, Dzmitry Bahdanau,
  Fethi Bougares, Holger Schwenk, and Yoshua Bengio. 2014.
\newblock \href {https://doi.org/10.3115/v1/D14-1179} {Learning phrase
  representations using {RNN} encoder{--}decoder for statistical machine
  translation}.
\newblock In \emph{Proceedings of the 2014 Conference on Empirical Methods in
  Natural Language Processing ({EMNLP} 2014)}, pages 1724--1734.

\bibitem[{Chorowski et~al.(2015)Chorowski, Bahdanau, Serdyuk, Cho, and
  Bengio}]{chorowski2015attention}
Jan~K Chorowski, Dzmitry Bahdanau, Dmitriy Serdyuk, Kyunghyun Cho, and Yoshua
  Bengio. 2015.
\newblock \href
  {https://papers.nips.cc/paper/5847-attention-based-models-for-speech-recognition.pdf}
  {Attention-based models for speech recognition}.
\newblock In \emph{Proceedings of the 29th Conference on Neural Information
  Processing Systems (NIPS 2015)}, pages 577--585.

\bibitem[{Cui et~al.(2019)Cui, Iida, Hung, Utsuro, and
  Nagata}]{cui-etal-2019-mixed}
Hongyi Cui, Shohei Iida, Po-Hsuan Hung, Takehito Utsuro, and Masaaki Nagata.
  2019.
\newblock \href {https://doi.org/10.18653/v1/D19-5622} {Mixed multi-head
  self-attention for neural machine translation}.
\newblock In \emph{Proceedings of the 3rd Workshop on Neural Generation and
  Translation}, pages 206--214.

\bibitem[{Dou et~al.(2019)Dou, Tu, Wang, Wang, Shi, and Zhang}]{dou2019dynamic}
Zi-Yi Dou, Zhaopeng Tu, Xing Wang, Longyue Wang, Shuming Shi, and Tong Zhang.
  2019.
\newblock \href
  {https://www.aaai.org/ojs/index.php/AAAI/article/view/3772/3650} {Dynamic
  layer aggregation for neural machine translation with routing-by-agreement}.
\newblock In \emph{Proceedings of the 33rd AAAI Conference on Artificial
  Intelligence (AAAI 2019)}, pages 86--93.

\bibitem[{Gehring et~al.(2017)Gehring, Auli, Grangier, Yarats, and
  Dauphin}]{gehring2017convolutional}
Jonas Gehring, Michael Auli, David Grangier, Denis Yarats, and Yann~N Dauphin.
  2017.
\newblock \href {https://dl.acm.org/citation.cfm?id=3305510} {Convolutional
  sequence to sequence learning}.
\newblock In \emph{Proceedings of the 34th International Conference on Machine
  Learning (ICML 2017)}, pages 1243--1252.

\bibitem[{Gong et~al.(2018)Gong, Qiu, Wang, and Huang}]{gong2018information}
Jingjing Gong, Xipeng Qiu, Shaojing Wang, and Xuanjing Huang. 2018.
\newblock \href {https://arxiv.org/abs/1806.01501} {Information aggregation via
  dynamic routing for sequence encoding}.
\newblock \emph{arXiv preprint arXiv:1806.01501}.

\bibitem[{Guo et~al.(2019)Guo, Qiu, Liu, Shao, Xue, and
  Zhang}]{guo-etal-2019-star}
Qipeng Guo, Xipeng Qiu, Pengfei Liu, Yunfan Shao, Xiangyang Xue, and Zheng
  Zhang. 2019.
\newblock \href {https://doi.org/10.18653/v1/N19-1133} {Star-transformer}.
\newblock In \emph{Proceedings of the 2019 Conference of the North {A}merican
  Chapter of the Association for Computational Linguistics: Human Language
  Technologies (NAACL 2019)}, pages 1315--1325.

\bibitem[{Hao et~al.(2019)Hao, Wang, Yang, Wang, Zhang, and
  Tu}]{hao-etal-2019-modeling}
Jie Hao, Xing Wang, Baosong Yang, Longyue Wang, Jinfeng Zhang, and Zhaopeng Tu.
  2019.
\newblock \href {https://doi.org/10.18653/v1/N19-1122} {Modeling recurrence for
  transformer}.
\newblock In \emph{Proceedings of the 2019 Conference of the North {A}merican
  Chapter of the Association for Computational Linguistics: Human Language
  Technologies (NAACL 2019)}, pages 1198--1207.

\bibitem[{Hassan et~al.(2018)Hassan, Aue, Chen, Chowdhary, Clark, Federmann,
  Huang, Junczys-Dowmunt, Lewis, Li et~al.}]{hassan2018achieving}
Hany Hassan, Anthony Aue, Chang Chen, Vishal Chowdhary, Jonathan Clark,
  Christian Federmann, Xuedong Huang, Marcin Junczys-Dowmunt, William Lewis,
  Mu~Li, et~al. 2018.
\newblock \href {https://arxiv.org/pdf/1803.05567.pdf} {Achieving human parity
  on automatic chinese to english news translation}.
\newblock \emph{arXiv preprint arXiv:1803.05567}.

\bibitem[{Hochreiter and Schmidhuber(1997)}]{hochreiter1997long}
Sepp Hochreiter and J{\"u}rgen Schmidhuber. 1997.
\newblock Long short-term memory.
\newblock \emph{Neural Computation}, 9(8):1735--1780.

\bibitem[{Klein et~al.(2017)Klein, Kim, Deng, Senellart, and
  Rush}]{klein-etal-2017-opennmt}
Guillaume Klein, Yoon Kim, Yuntian Deng, Jean Senellart, and Alexander Rush.
  2017.
\newblock \href {https://www.aclweb.org/anthology/P17-4012} {{O}pen{NMT}:
  Open-source toolkit for neural machine translation}.
\newblock In \emph{Proceedings of {ACL} 2017, System Demonstrations}, pages
  67--72.

\bibitem[{Li et~al.(2019)Li, Yang, Dou, Wang, Lyu, and
  Tu}]{li-etal-2019-information}
Jian Li, Baosong Yang, Zi-Yi Dou, Xing Wang, Michael~R. Lyu, and Zhaopeng Tu.
  2019.
\newblock \href {https://doi.org/10.18653/v1/N19-1359} {Information aggregation
  for multi-head attention with routing-by-agreement}.
\newblock In \emph{Proceedings of the 2019 Conference of the North {A}merican
  Chapter of the Association for Computational Linguistics: Human Language
  Technologies (NAACL 2019)}, pages 3566--3575.

\bibitem[{Lin et~al.(2017)Lin, Feng, Santos, Yu, Xiang, Zhou, and
  Bengio}]{lin2017structured}
Zhouhan Lin, Minwei Feng, Cicero Nogueira~dos Santos, Mo~Yu, Bing Xiang, Bowen
  Zhou, and Yoshua Bengio. 2017.
\newblock \href {https://arxiv.org/pdf/1703.03130.pdf} {A structured
  self-attentive sentence embedding}.
\newblock In \emph{Proceedings of the 5th International Conference on Learning
  Representations (ICLR 2017)}.

\bibitem[{Liu et~al.(2019)Liu, Lin, Liu, Xu, Ren, Diao, and
  Yang}]{liu-etal-2019-transformer}
Jintao Liu, Hongfei Lin, Xikai Liu, Bo~Xu, Yuqi Ren, Yufeng Diao, and Liang
  Yang. 2019.
\newblock \href {https://www.aclweb.org/anthology/W19-5511} {Transformer-based
  capsule network for stock movement prediction}.
\newblock In \emph{Proceedings of the First Workshop on Financial Technology
  and Natural Language Processing}, pages 66--73.

\bibitem[{Luong et~al.(2015)Luong, Pham, and
  Manning}]{luong-etal-2015-effective}
Thang Luong, Hieu Pham, and Christopher~D. Manning. 2015.
\newblock \href {https://doi.org/10.18653/v1/D15-1166} {Effective approaches to
  attention-based neural machine translation}.
\newblock In \emph{Proceedings of the 2015 Conference on Empirical Methods in
  Natural Language Processing (EMNLP 2015)}, pages 1412--1421.

\bibitem[{Papineni et~al.(2002)Papineni, Roukos, Ward, and Zhu}]{bleu}
Kishore Papineni, Salim Roukos, Todd Ward, and Wei-Jing Zhu. 2002.
\newblock \href {https://doi.org/10.3115/1073083.1073135} {{B}leu: a method for
  automatic evaluation of machine translation}.
\newblock In \emph{Proceedings of the 40th Annual Meeting of the Association
  for Computational Linguistics (ACL 2002)}, pages 311--318.

\bibitem[{Peters et~al.(2018)Peters, Neumann, Iyyer, Gardner, Clark, Lee, and
  Zettlemoyer}]{peters-etal-2018-deep}
Matthew Peters, Mark Neumann, Mohit Iyyer, Matt Gardner, Christopher Clark,
  Kenton Lee, and Luke Zettlemoyer. 2018.
\newblock \href {https://doi.org/10.18653/v1/N18-1202} {Deep contextualized
  word representations}.
\newblock In \emph{Proceedings of the 2018 Conference of the North {A}merican
  Chapter of the Association for Computational Linguistics: Human Language
  Technologies (NAACL 2018)}, pages 2227--2237.

\bibitem[{Raganato and Tiedemann(2018)}]{raganato-tiedemann-2018-analysis}
Alessandro Raganato and J{\"o}rg Tiedemann. 2018.
\newblock \href {https://doi.org/10.18653/v1/W18-5431} {An analysis of encoder
  representations in transformer-based machine translation}.
\newblock In \emph{Proceedings of the 2018 {EMNLP} Workshop {B}lackbox{NLP}:
  Analyzing and Interpreting Neural Networks for {NLP}}, pages 287--297.

\bibitem[{Sabour et~al.(2017)Sabour, Frosst, and Hinton}]{sabour2017dynamic}
Sara Sabour, Nicholas Frosst, and Geoffrey~E Hinton. 2017.
\newblock \href
  {https://papers.nips.cc/paper/6975-dynamic-routing-between-capsules.pdf}
  {Dynamic routing between capsules}.
\newblock In \emph{Proceedings of the 31th Conference on Neural Information
  Processing Systems (NIPS 2017)}, pages 3856--3866.

\bibitem[{Sennrich et~al.(2016)Sennrich, Haddow, and
  Birch}]{sennrich2016neural}
Rico Sennrich, Barry Haddow, and Alexandra Birch. 2016.
\newblock \href {https://doi.org/10.18653/v1/P16-1162} {Neural machine
  translation of rare words with subword units}.
\newblock In \emph{Proceedings of the 54th Annual Meeting of the Association
  for Computational Linguistics (ACL 2016)}, pages 1715--1725.

\bibitem[{Shaw et~al.(2018)Shaw, Uszkoreit, and Vaswani}]{shaw-etal-2018-self}
Peter Shaw, Jakob Uszkoreit, and Ashish Vaswani. 2018.
\newblock \href {https://doi.org/10.18653/v1/N18-2074} {Self-attention with
  relative position representations}.
\newblock In \emph{Proceedings of the 2018 Conference of the North {A}merican
  Chapter of the Association for Computational Linguistics: Human Language
  Technologies (NAACL 2018)}, pages 464--468.

\bibitem[{Shen et~al.(2018)Shen, Zhou, Long, Jiang, Pan, and
  Zhang}]{shen2018disan}
Tao Shen, Tianyi Zhou, Guodong Long, Jing Jiang, Shirui Pan, and Chengqi Zhang.
  2018.
\newblock \href
  {https://www.aaai.org/ocs/index.php/AAAI/AAAI18/paper/viewFile/16126/16099}
  {Di{SAN}: Directional self-attention network for {RNN}/{CNN}-free language
  understanding}.
\newblock In \emph{Proceedings of the 32nd AAAI Conference on Artificial
  Intelligence (AAAI 2018)}, pages 5446--5455.

\bibitem[{Vaswani et~al.(2017)Vaswani, Shazeer, Parmar, Uszkoreit, Jones,
  Gomez, Kaiser, and Polosukhin}]{vaswani2017attention}
Ashish Vaswani, Noam Shazeer, Niki Parmar, Jakob Uszkoreit, Llion Jones,
  Aidan~N Gomez, {\L}ukasz Kaiser, and Illia Polosukhin. 2017.
\newblock \href
  {https://papers.nips.cc/paper/7181-attention-is-all-you-need.pdf} {Attention
  is all you need}.
\newblock In \emph{Proceedings of the 31th Conference on Neural Information
  Processing Systems (NIPS 2017)}, pages 5998--6008.

\bibitem[{Wang et~al.(2019)Wang, Lee, and Chen}]{wang-etal-2019-tree}
Yaushian Wang, Hung-Yi Lee, and Yun-Nung Chen. 2019.
\newblock \href {https://doi.org/10.18653/v1/D19-1098} {Tree transformer:
  Integrating tree structures into self-attention}.
\newblock In \emph{Proceedings of the 2019 Conference on Empirical Methods in
  Natural Language Processing and the 9th International Joint Conference on
  Natural Language Processing (EMNLP-IJCNLP)}, pages 1061--1070.

\bibitem[{Wu et~al.(2016)Wu, Schuster, Chen, Le, Norouzi, Macherey, Krikun,
  Cao, Gao, Macherey et~al.}]{wu2016google}
Yonghui Wu, Mike Schuster, Zhifeng Chen, Quoc~V Le, Mohammad Norouzi, Wolfgang
  Macherey, Maxim Krikun, Yuan Cao, Qin Gao, Klaus Macherey, et~al. 2016.
\newblock \href {https://arxiv.org/pdf/1609.08144.pdf} {Google's neural machine
  translation system: Bridging the gap between human and machine translation}.
\newblock \emph{arXiv preprint arXiv:1609.08144}.

\bibitem[{Xiao et~al.(2019)Xiao, Li, Zhao, Wang, and
  Chen}]{xiao-etal-2019-lattice}
Fengshun Xiao, Jiangtong Li, Hai Zhao, Rui Wang, and Kehai Chen. 2019.
\newblock \href {https://doi.org/10.18653/v1/P19-1298} {Lattice-based
  transformer encoder for neural machine translation}.
\newblock In \emph{Proceedings of the 57th Annual Meeting of the Association
  for Computational Linguistics}, pages 3090--3097.

\bibitem[{Yang et~al.(2019{\natexlab{a}})Yang, Li, Wong, Chao, Wang, and
  Tu}]{yang2019context}
Baosong Yang, Jian Li, Derek~F Wong, Lidia~S Chao, Xing Wang, and Zhaopeng Tu.
  2019{\natexlab{a}}.
\newblock \href
  {https://www.aaai.org/ojs/index.php/AAAI/article/view/3809/3687}
  {Context-aware self-attention networks}.
\newblock In \emph{Proceedings of the 33rd AAAI Conference on Artificial
  Intelligence (AAAI 2019)}, pages 387--394.

\bibitem[{Yang et~al.(2018{\natexlab{a}})Yang, Tu, Wong, Meng, Chao, and
  Zhang}]{yang-etal-2018-modeling}
Baosong Yang, Zhaopeng Tu, Derek~F. Wong, Fandong Meng, Lidia~S. Chao, and Tong
  Zhang. 2018{\natexlab{a}}.
\newblock \href {https://www.aclweb.org/anthology/D18-1475} {Modeling localness
  for self-attention networks}.
\newblock In \emph{Proceedings of the 2018 Conference on Empirical Methods in
  Natural Language Processing (EMNLP 2018)}, pages 4449--4458.

\bibitem[{Yang et~al.(2019{\natexlab{b}})Yang, Wang, Wong, Chao, and
  Tu}]{yang-etal-2019-convolutional}
Baosong Yang, Longyue Wang, Derek~F. Wong, Lidia~S. Chao, and Zhaopeng Tu.
  2019{\natexlab{b}}.
\newblock \href {https://www.aclweb.org/anthology/N19-1407} {Convolutional
  self-attention networks}.
\newblock In \emph{Proceedings of the 2019 Conference of the North {A}merican
  Chapter of the Association for Computational Linguistics: Human Language
  Technologies (NAACL 2019)}, pages 4040--4045.

\bibitem[{Yang et~al.(2018{\natexlab{b}})Yang, Zhao, Ye, Lei, Zhao, and
  Zhang}]{yang-etal-2018-investigating}
Min Yang, Wei Zhao, Jianbo Ye, Zeyang Lei, Zhou Zhao, and Soufei Zhang.
  2018{\natexlab{b}}.
\newblock \href {https://doi.org/10.18653/v1/D18-1350} {Investigating capsule
  networks with dynamic routing for text classification}.
\newblock In \emph{Proceedings of the 2018 Conference on Empirical Methods in
  Natural Language Processing (EMNLP 2018)}, pages 3110--3119.

\bibitem[{Yang et~al.(2016)Yang, Yang, Dyer, He, Smola, and
  Hovy}]{yang-etal-2016-hierarchical}
Zichao Yang, Diyi Yang, Chris Dyer, Xiaodong He, Alex Smola, and Eduard Hovy.
  2016.
\newblock \href {https://doi.org/10.18653/v1/N16-1174} {Hierarchical attention
  networks for document classification}.
\newblock In \emph{Proceedings of the 2016 Conference of the North {A}merican
  Chapter of the Association for Computational Linguistics: Human Language
  Technologies (NAACL 2016)}, pages 1480--1489.

\end{thebibliography}
\bibliographystyle{acl_natbib}

\end{document}